**Embodied AI in Machine Learning – is it Really Embodied?**

Matej Hoffmann and Shubhan Parag Patni
Department of Cybernetics, Faculty of Electrical Engineering, Czech Technical University in Prague

**Introduction**

Embodied Artificial Intelligence (Embodied AI) is gaining momentum in the machine learning communities with the goal of leveraging current progress in AI (deep learning, transformers, large language and visual-language models) to empower robots. In this chapter we put this work in the context of "Good Old-Fashioned Artificial Intelligence" (GOFAI) (Haugeland, 1989) and the behavior-based or embodied alternatives (R. A. Brooks 1991; Pfeifer and Scheier 2001). We claim that the AI-powered robots are only weakly embodied and inherit some of the problems of GOFAI. Moreover, we review and critically discuss the possibility of cross-embodiment learning (Padalkar et al. 2024). We identify fundamental roadblocks and propose directions on how to make progress.

*GOFAI-powered robots never really worked*
Artificial Intelligence (AI) in the 1950's and 1960's, later called "Good Old-Fashioned Artificial Intelligence" (GOFAI) by Haugeland(1989), held that the key to intelligence is computation with symbols that represent the world. The keywords were algorithmic nature, symbolic computation and representation. GOFAI was very successful in formal domains (like chess), where the state of the world is discrete and directly accessible and standard AI techniques (like search) can be applied. While the focus has been on abstract "thinking", when entering the real world, a relationship had to be established between the dynamic, continuous, partially accessible reality out there and the internal world representation. That is, reality had to be sensed and mapped onto the internal world model, in which the "thinking" was performed. Finally, whatever action was selected, it had to be executed in the real world. The approach thus became known as the *sense-think-act* architecture. The "interfaces" with the real world—previously uninteresting and underestimated—became fundamental and practical challenges. The "frame problem"' (keeping the internal representation of the world consistent with the real world outside) and the "symbol-grounding problem"' (concerned with the relationship of the symbolic representation and the outside world) are the most serious of the fundamental problems. The initial hopes of applying for example theorem solving (e.g., (Fikes and Nilsson 1971)) to real-world robot tasks have never really materialized and AI and robotics have developed largely independently, meeting when complex high-level plans and hence reasoning were required (Russell and Norvig 2021; Beetz et al. 2016).[1]

*Robotics before the learning era*
It is not within the scope of this chapter to provide an overview of the field of robotics and its history. Before the "robot learning era" in recent years, robotics research focused on developing the foundations (kinematics, dynamics, motion planning and control), sensing and perception, localization and mapping, as well as industrial robotics and, more recently, human-robot interaction (see the corresponding parts in (Siciliano and Khatib 2016)). Robotics had more intersections with control theory than with AI. The impressive robot agility as demonstrated by the robots from Boston Dynamics like Atlas have been achieved through careful engineering and the use of model predictive control rather than learning.

---

[1] In (Russell and Norvig 2021), there is only one Chapter (26) on Robotics. Conversely, In (Siciliano and Khatib 2016), there is only one chapter (Beetz et al. 2016) on AI reasoning methods for robotics.

*Embodiment, behavior-based, and soft robotics*
During the GOFAI era, the computer was not only a tool but also a metaphor for the mind in the paradigm known as cognitivism (Fodor 1975). The response to this in cognitive science was embodied cognition which holds that cognitive processes are constitutively shaped by the interaction with the world through the agent's body, emphasizing the role of action rather than world modeling and representations (e.g., (Engel et al. 2013)). In order to demonstrate that behaviors, which would be considered intelligent or cognitive by many, do not have to come from internal world models and computation over them, machines with minimalistic controllers were built and their interaction with the environment was observed. Grey Walter was the pioneer of this approach, building electronic machines with a minimal "brain" that displayed phototactic-like behavior (Walter 1953). This was picked up by Valentino Braitenberg (1986) who built a series of two-wheeled vehicles of increasing complexity. Already the most primitive ones, in which sensors were directly connected to motors (exciting or inhibiting them), displayed sophisticated behaviors. Although the driving mechanisms are simple and entirely deterministic, the interaction with the real world gives rise to complex behavioral patterns.

Motivated by practical considerations, i.e. to build robots that can interact with the world in real time, Rodney Brooks openly attacked the GOFAI position in the seminal article "Intelligence without representation" (R. A. Brooks 1991). Through building robots that interact with the real world, such as insect robots, he realized that "when we examine very simple level intelligence we find that explicit representations and models of the world simply get in the way. It turns out to be better to use the world as its own model." Inspired by biological evolution, Brooks created a decentralized control architecture consisting of different layers; every layer is a more or less simple coupling of sensors to motors. The levels operate in parallel, but are built in a hierarchy (hence the term subsumption architecture; (R. Brooks 1986)).

Embodiment means more than only "intelligence requires a body". The behavior of any system is not merely the outcome of an internal control structure (such as the central nervous system); it is also affected by the ecological niche in which the system is physically embedded, by its morphology (the shape of its body and limbs, as well as the type and placement of sensors and effectors), and by the material properties of the elements composing the morphology (Pfeifer and Bongard 2006). Taking embodiment seriously is mandatory for the rapidly developing field of soft robotics (e.g., (Yasa et al. 2023)) where the complexity and nonlinearity of the robot bodies makes it hard to employ traditional control schemes and the dynamics of the material substrate needs to be exploited.

*Current Weakly Embodied AI (WEAI)*
Large-scale, so-called foundation models, in Natural Language Processing (NLP) and computer vision can enable capable AI systems by providing general-purpose pretrained models which can often outperform their narrowly targeted counterparts trained on smaller but more task-specific data. Applying the same strategy to control robots is appealing. The community originating in computer vision, machine learning, and NLP, but now connecting their models to robots uses the label "Embodied AI" (e.g., (Deitke et al. 2022; Liu et al. 2024; Vanhoucke 2024)). The idea is to leverage the reasoning ("common sense") capabilities of such models as well as their capability to understand visual scenes (images) in order to produce plans that a robot can execute.

In this chapter, we argue that the embodiment in these works is weak or shallow and discuss the implications this has on the field—what are principled roadblocks and where are productive future research directions.

**Implications of embodiment for making better robots**
The implications of embodiment for any agent (biological or artificial) are illustrated graphically in Fig. 1 (Pfeifer, Lungarella, and Iida 2007).

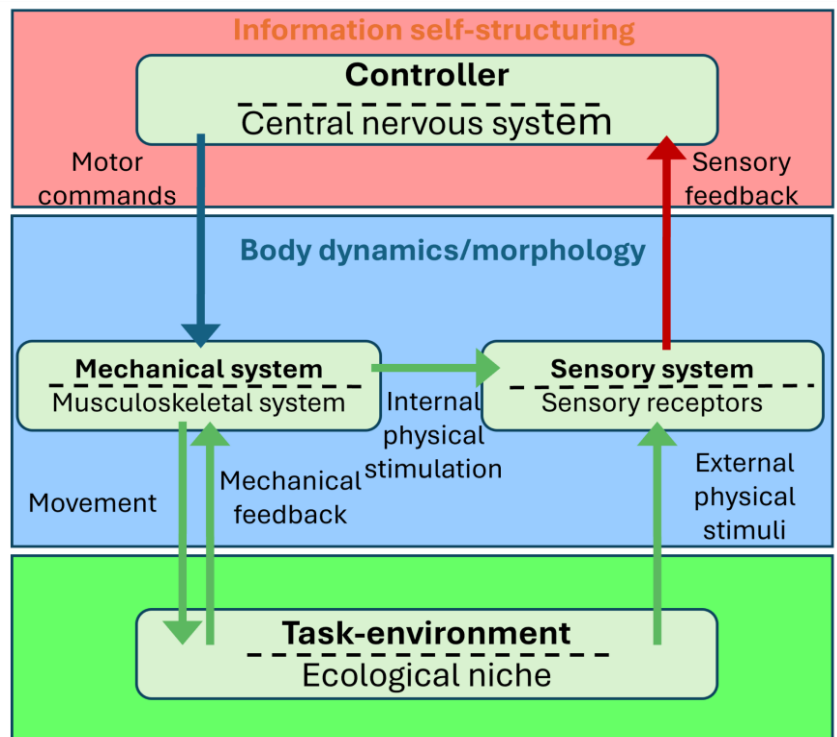

Figure 1. Implications of embodiment (the interplay of information and physical processes). Driven by motor commands, the musculoskeletal system (mechanical system) of the agent acts on the external environment (task environment or ecological niche). The action leads to rapid mechanical feedback characterized by pressure on the bones, torques in the joints, and passive deformation of skin tissue. In parallel, external stimuli (pressure, temperature, and electromagnetic fields) and internal physical stimuli (forces and torques developed in the muscles and joint-supporting ligaments, as well as accelerations) impinge on the sensory receptors (sensory system). The patterns induced thus depend on the physical characteristics and morphology of the sensory systems and on the motor commands. Especially if the interaction is sensory-motor coordinated, as in foveation, reaching, or grasping movements, information structure is generated. The effect of the motor command strongly depends on the tunable morphological and material properties of the musculoskeletal system, where by tunable we mean that properties such as shape and compliance can be changed dynamically: During the forward swing of the leg in walking, the muscles should be largely passive, whereas when hitting the ground, high stiffness is required, so that the materials can take over some of the adaptive functionality on impact, such as the damped oscillation of the knee joint.  Figure and caption from (Pfeifer, Lungarella, and Iida 2007)(figure has been redrawn).

Below we list selected premises of embodiment to serve as a checklist to gauge how embodied WEAI really is. The *body* or body *morphology* refers to the shape of the body and limbs, as well as the type and placement of sensors and effectors. The list draws on the principles for intelligent systems from (Pfeifer and Scheier 2001) and (Pfeifer and Bongard 2006, chap. 4).

**Behavior is not in the brain/controller but in the closed-loop interaction of the controller, the body, and the environment.**
1. **Body morphology facilitates control.** Interaction of the body with the environment can facilitate the production of behavior with simpler control by the nervous system. In extreme cases, only physics, i.e. no sensing, control, or even actuation may suffice to produce behavior like walking (McGeer 1990). Mechanical feedback loops (see the movement and mechanical feedback arrows in Fig. 1) can even serve to stabilize the system (see self-stabilization in (Pfeifer, Lungarella, and Iida 2007) or physical implications of embodiment in (Hoffmann and Pfeifer 2012)). Complex nonlinear bodies offer a bigger potential for their intrinsic dynamics to be exploited to produce desired behavior as studied in soft robotics (Hauser, Nanayakkara, and Forni 2023).
2. **Sensor morphology facilitates perception.** The placement, shape and mechanical properties of sensors importantly determine what will be sensed. This is certainly true for tactile sensors, for example, but also for visual sensors (Franceschini et al. 1992). In biology, the eyes of different animals have evolved to facilitate perception. In the realm of insect eyes, motion detection, which can be used for obstacle avoidance, can often be simplified if the light-sensitive cells are not spaced evenly. In the compound eye of the house fly, the spacing of the facets is denser toward the front of the animal, which compensates for the phenomenon of motion parallax, i.e. the fact that at constant speed, objects on the side travel faster across the visual field than objects towards the front. Allowing for some idealization, this implies that under the condition of straight flight, the same motion detection circuitry—the elementary motion detectors—can be employed for motion detection for the entire eye, a principle that has also been applied to the construction of navigating robots (Hoshino, Mura, and Shimoyama 2000).
3. **Sensorimotor coordination and active perception.** Closed-loop interaction with the environment and the choice of specific actions may significantly transform what is perceived and make a perceptual task like discrimination much easier compared to passive perception or random movements of the agent. Such *information self-structuring* in simple robots was quantified in Lungarella and Sporns(2006) using information theoretic measures. The works on active perception (in particular active vision) (Bajcsy, Aloimonos, and Tsotsos 2018) and the principle of sensorimotor coordination (Pfeifer and Bongard 2006) are relevant here.
4. **Parallel loosely coupled processes.** Intelligence—at least as we know it from biology—is emergent from a large number of parallel processes[2] coordinated through the embodied interaction with the environment (Pfeifer and Bongard 2006, chap. 4). The subsumption architecture (R. Brooks 1986) makes a similar point but adds a hierarchy to the parallel processes.
5. **Principle of ecological balance.**
    a. Given a certain task environment, there has to be a match between the complexities of the agent's sensory, motor, and neural systems.
    b. There is a certain balance or task distribution between morphology, materials, control, and environment.

Embodiment somewhat overlaps with the notion of *morphological computation* but we want to emphasize the role of the body for facilitating behavior and hence simplifying control and for perception. We do not regard these processes as computational (in line with (Müller and Hoffmann 2017)).

---

[2] These are parallel interaction loops—mechanical or informational—operating at different time scales. It has nothing to do with the possibility of parallelization on a Graphical Processing Unit (GPU).

**Representative works of Weakly Embodied AI**
A Generalist Agent (GATO) (Reed et al. 2022) can be regarded as a natural extension of the machine learning and NLP models, as it retains the capabilities of the image- and language-based foundation models like image caption generation and chat, combines it with playing Atari games, and also brings a real robotics task into the same model—the RGB Stacking benchmark, where a robot arm is stacking coloured blocks using input from an RGB camera.

There is an explosion of works in "Embodied AI" (here WEAI), i.e. the use of large models in robotics. To give the discussion concrete contours, we briefly describe the evolution of this young field (we discuss works from 2022 to 2024) inspired by the perspective of (Vanhoucke 2024). The deployment of Large Language Models (LLMs) into robotics has naturally started from planning, which was lifted into "semantic space" from geometric space (see Fig. 2A). An example is the "Say-Can" model (Ahn et al. 2022). The reasoning and planning capabilities traditionally handled by automatic inference systems in the GOFAI era now leverage the "common sense" power of LLMs.

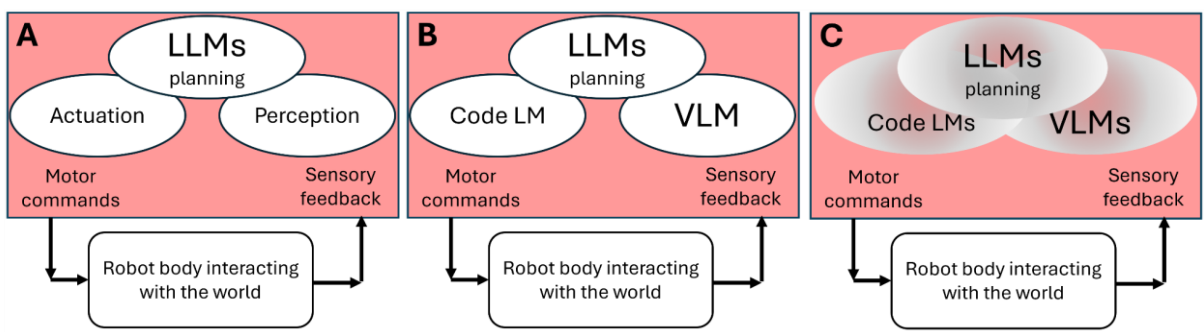

Figure 2. Evolution of approaches in Embodied AI. Redrawn and adapted from (Vanhoucke 2024).

The next modules to be "AI-powered" have been the Perception and Actuation modules, through Visual Language Models (VLMs) and "Code LMs", respectively, as shown in Fig. 2B.

The world state, now represented in language, had to be extracted from the world (perception or state estimation module). The next logical step thus was to deploy Visual Language Models (VLMs), which can convert images into language representations, as the Perception module (e.g., Socrates, (Zeng et al. 2022)). To keep the representation in sync with the real world (addressing the frame problem), periodic state reestimation and replanning is necessary, as exemplified in the "Inner Monologue" (Huang et al. 2022).

The final bastion of traditional robotics that was given LLM treatment (in the words of (Vanhoucke 2024)) was the action. This was achieved by having the language model generate code to be executed by the robot ("Code LM"), as in "Code as Policies" (Liang et al. 2023), Text2Motion (Lin et al. 2023), ProgPrompt (Singh et al. 2023), or ChatGPT for Robotics (Vemprala et al. 2024).

The interfaces between the modules—LLM as a planner, VLM as state estimator, and Code LM as action generator—have become a bottleneck of these architectures. Therefore, the next step was to "blend these modules in" as schematically illustrated in Fig 2C. PALM-E (Driess et al. 2023) blended the LLM and VLM; RT-1 (Brohan, Brown, Carbajal, Chebotar, Dabis, et al. 2023) blended the Perception and Actuation (VLM and Code LM). The final logical next step was to reason jointly about the entire problem, blending all the module boundaries while still leveraging "internet data". Examples of this approach are RT-2 (Brohan, Brown, Carbajal, Chebotar, Chen, et al. 2023) or VC-1 (Yokoyama et al. 2023).

**How embodied is Weakly Embodied AI?**
Comparing Fig. 1 and Fig. 2, it is apparent that WEAI is really weakly embodied. The implications of embodiment in Fig. 1 are centered on body dynamics (blue) and task-environment (green) blocks. In the WEAI architectures (Fig. 2), what matters is the controller or brain and the interaction with the world through the agent body is reduced to a mere arrow closing the loop between the sensors and actuators.
Let us look through the "embodiment checklist" and see where WEAI stands.
   0. Behavior is not in the brain/controller but in the closed-loop interaction of the controller, the body, and the environment.
      - In WEAI, the behavior is in the controller—in the large model. Complete long-horizon tasks can be planned in the model and the state in the model has to be kept in sync with the world. Embodied interaction with the environment is not exploited.
   1. Body morphology facilitates control.
      - In WEAI, there is little room for the agent body to be exploited to fulfill the task. The robot platforms used—like mobile robots or robot arms with 2-finger grippers—do not feature rich body dynamics that could be exploited for the task. Moreover, the details of the sensory and motor interfaces of the robots are abstracted away. For example, if the action space is the position and orientation of the end effector and the open/close gripper command, it leaves little room for the model to make interesting use of the robot morphology.
   2. Sensor morphology facilitates perception.
      - WEAI leverages the power of deep neural networks to find patterns in static images (e.g., object recognition) and how what is on the images can be translated into text (VLMs). The inputs need to be more or less standard RGB images. The power of Convolutional Neural Networks (CNNs) and Visual Transformers on tasks like recognition will hardly transfer to particular sensor morphologies like with non-uniform spacing of light sensitive cells. Thus, sensor morphologies designed for a particular task environment are not compatible with the WEAI approach.
   3. Sensorimotor coordination and active perception.
      - Typical WEAI architectures aim to maximally exploit the "common sense" of LLMs to solve tasks without the need for additional training—also called "zero shot". In this case, there is no possibility to exploit the closed-loop embodied interaction with the environment. When additional training (finetuning) of the model takes place, it typically uses offline datasets collected when for example a human was teleoperating the robot to solve the task. The operator may exploit sensorimotor coordination to solve the task and this would be transferred to the

robot controller. However, there has to be a match in the *situatedness*[3], i.e. the human and the robot will have to see through similar "eyes". If the operator drives the robot using for example a third person view but the robot has an egocentric camera, the transfer will not work.
4. Parallel loosely coupled processes.
    ○ WEAI architectures may consist of several blocks—VLM → LLM → Code LM—with parallel processing inside, but there is typically a single loop connecting the sensors and motors. Hence, there is only one process and one time scale.
5. Principle of ecological balance.
    ○ WEAI is strongly susceptible to ecological imbalance as the architectures feature a gigantic brain. The sensory inputs can be text and vision, with some developments in adding other modalities. The robots typically have rather simple morphology and hence action possibilities. However, the main source of ecological imbalance is the action representation—typically a low-dimensional discrete action space.

**Active embodied interaction versus offline learning**
The impressive progress in NLP and computer vision has been enabled by two main factors: (1) large datasets of text and images were available; (2) passive supervised learning was appropriate for tasks like predicting the next word in a sequence or labeling the object in the image. A big part of the WEAI community thinks that getting enough data will "solve robotics" as well. Datasets are being collected typically using teleoperation. From such datasets, imitation learning or offline reinforcement learning methods are applied to train a controller (e.g., (Walke et al. 2023)). However, successful behavior in the real world inherently requires sensorimotor coordination and active perception (see also a recent article on the limits of passive AI (Pezzulo et al. 2024)). If the robot cannot sample the actions and their consequences itself, learning is bound to be inefficient. We want to illustrate this with a cartoon in Fig. 3. The left panel (A) shows the setup from the seminal work of Held and Hein (1963). Kittens were raised in darkness from birth until they entered the experimental setup. The kitten from the first group (on the right in Fig. 3A) was allowed to actively move in the arena, while the kitten from the second group was passively moved around on a trolley, driven by the movement of the first kitten. After this "training phase", kittens from both groups were presented with behavioral tests (visually-guided paw placement, avoidance of visual cliff, blink to an approaching object). Importantly, only the kittens from the actively moving group succeeded in the tests of visually guided behavior. The right panel (B) is a cartoon-like representation of how robot datasets are created. A human (on the right) is remotely controlling a robot (a car equipped with sensors in this case) and the data is recorded. The robot has one important advantage. While the passive kitten did not have access to the motor commands of the active kitten, the robot dataset collected may contain both sensory and motor signals (if the operator uses the same actuation interface). However, the robot still does not have the possibility to actively sample the actions and see their effects and associated rewards, as in standard online reinforcement learning.

---

[3] "An agent is "situated" if it can acquire information about the current situation through its sensors in interaction with the environment." (Pfeifer and Scheier 2001, 72)

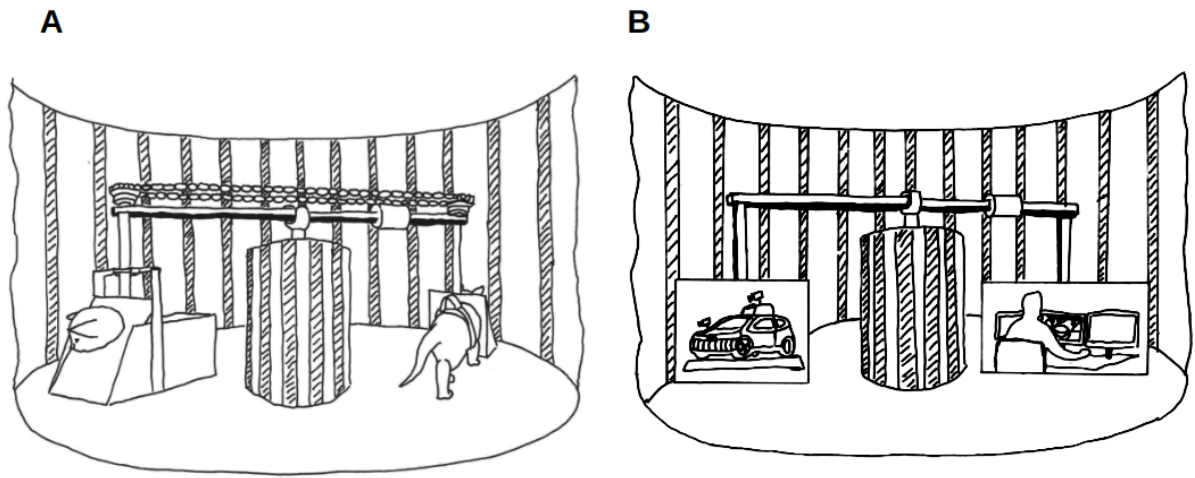

Figure 3. Active embodied interaction versus being driven around. (A) Setup from (Held and Hein 1963) (redrawn). The kitten on the right is actively walking in the apparatus; the kitten on the left is being passively carried around but receiving similar stimulation. (B) A cartoon-like representation dataset generation in WEAI. A human teleoperates a robot which only passively receives sensory stimulation. (Panel A from (Held and Hein 1963))

**Natural WEAI extensions**
WEAI researchers are aware of the gap that still exists between "high-level language knowledge" and "low-level robotics knowledge" and there is work under way to bridge it. For example, there are several extensions to the RT family of models that bring motion-centric representations (RT-Trajectory (Gu et al. 2023)), prompting with visual iterative optimization (PIVOT – Iterative visual prompting (Nasiriany et al. 2024)), or introduce hierarchies to increase the granularity of the language instructions (e.g., for "put the coke on the table", there would be another more granular description starting with "move arm forward", etc.) (RT-Hierarchy (Belkhale et al. 2024) ). Szot et al. (2024) developed a multimodal LLM and an action space adapter for continuous action spaces. Touch is a sensory modality that is especially relevant for manipulation and that researchers are trying to add to the models (e.g., (Sferrazza et al. 2023; F. Yang et al. 2024)).

**Cross-embodiment Embodied AI – an oxymoron?**
The need for large datasets to train the models triggered several cross-embodiment initiatives, that is "crowd-sourcing" data from different labs and different robots performing various tasks in order to train models that can master the different tasks, environment, and embodiments (see (Jaquier et al. 2024) for a review of transfer learning in robotics).

Open X-Embodiment is a large initiative across many institutions focusing primarily on developing a large robot dataset (Padalkar et al. 2024). Currently, 60 datasets have been brought into a common format, featuring 22 different robot embodiments (mostly robot arms). Building on previous models, RT-1 and RT-2, the new model, RT-X, receives a history of recent RGB images from a camera placed over the robot's shoulder, and language instructions (e.g., "pick up the orange fruit") as input and produces (predicts) a 7-dimensional action vector controlling the end effector x, y, z, roll, pitch, yaw, and gripper opening or the rates of these quantities. The action is tokenized into 256 bins uniformly distributed along each dimension. The results show some positive transfer such that co-training on data collected on multiple robots improves performance on a training task. However, generalization to new robots was not yet studied.

J. Yang et al. (2024) have put together a dataset featuring a heterogeneous set of robots (cars, quadrupedal robots, wheeled robots, and manipulator arms), aggregated from 18 manipulation, navigation, and driving dataset. Here, no language instructions are part of the model. The input to the network is a history of RGB images and the goal expressed as an egocentric image (object being grasped by the manipulator; mobile robot reaching a waypoint). The output is similar to the RT-X model: a 7-dimensional discrete action space. A single goal-conditioned policy was trained, capable of controlling robotic arms, quadcopters, quadrupeds, and mobile bases. They found that co-training with navigation data can enhance robustness and performance in goal-conditioned manipulation with a wrist-mounted camera. Zero-shot generalization to a new embodiment was evaluated.

"All Robots in One" (Wang et al. 2024) is another cross-embodiment initiative focusing solely on acquiring a dataset and extending the typical sensory inputs (text and vision) with audio and touch.

At first glance, embodied AI across embodiments seems contradictory. A key implication of embodied cognition is that our bodies constitutively shape the way we think (Pfeifer and Bongard 2006) and hence different bodies require different brains. How can a single large model (or "brain") command bodies as diverse as cars, legged robots, and robot arms?

The reason why the efforts reviewed above have had some success is that the *embodiment there is shallow*. The works have essentially created an abstraction layer that reduces every robot to the Cartesian coordinates of its gripper or the coordinates of the robot's center of mass in case of a mobile robot. The models were trained to learn a mapping from camera images to a low-dimensional action space, conditioned on a goal expressed in language ("pick apple from ... and place on …" (Padalkar et al. 2024) or as a goal image (J. Yang et al. 2024). The fact that the robot has joints—a kind of defining feature of a robot arm—is abstracted away. The control has a coarse spatial (discretized low-dim. action space) and temporal resolution (3-10 Hz). Interaction forces are not considered. Thus, one could easily conclude that this is not "real robotics" or "real embodiment".

**Fundamental roadblocks – do foundation models stand in the way?**

*Where WEAI is not like GOFAI*
While GOFAI and cognitivism rested on the manipulation of symbols that represent the world, WEAI could be classified as emergentism or connectionism, harnessing the power of neural networks.[4] WEAI representations are learned from data and are much more flexible compared to the largely handcrafted rule-based systems. Large models can also be more flexibly connected to the real world through VLMs for perception (mitigating the frame problem) and Code LMs for action. Finally, learning can be employed throughout the whole pipeline—end-to-end when appropriate.

*Where WEAI inherits GOFAI problems*
1. Symbol grounding problem. LLMs are praised for having "common sense" which can then be leveraged not only to chat but also to make action plans for the real world. This common sense comes from the large text corpora that the models have digested. This provides much better grounding compared to GOFAI. In fact, Harnad (1990) proposes connectionism as a possible remedy to the symbol grounding problem. However, grounding derived solely from text corpora has intrinsic limitations. An LLM may hold a conversation on riding a bicycle, skiing, or fishing, but proper grounding on performing this in the real world when embodied in a robot is largely lacking. If language is the "common currency" and the LLM modules are used for reasoning, the problem of grounding is inevitable. Human cognition is grounded (Barsalou 2008) through internal multimodal sensorimotor simulations. Using Multimodal Large Language Models (MLLM) and grounding them into different embodiments and their associated action spaces will be necessary (Szot et al. 2024).
2. Embodiment and situatedness. In WEAI, intelligence is in the LLM (or MLLM). Knowledge extracted from texts, images and videos constitutes a *world model*, which is largely disembodied or indirectly shallowly embodied in a human-like embodiment (since the text and images on the Internet are human-centered). As the details of the body morphology are not represented, they cannot be meaningfully exploited by the "reasoning engine" for facilitating control or perception. The large model also somehow stands in the way of exploiting sensorimotor coordination or direct perception (Gibson 1979)—shortcuts exploiting closed-loop interaction with the environment.
3. "Soft real-time" interaction. A long time has passed since the Stanford Cart of Hans Moravec where long "thinking" periods separated every movement (Moravec 1983). Now we have much more computational power but also the models have grown considerably. Closing the loop at 3-10 Hz (Padalkar et al. 2024) is not fast enough for real tasks in robotics.

---
[4] The fact that LLMs use tokens, which can also be regarded as symbols, does not play an important part here.

*Where WEAI brings its own new challenges*
The large models are truly large, bringing enormous computational and energy costs. Even inference, that is only "running" the model, requires considerable computational resources. For example, training GPT-3 (now a "small and old" model) on only text data required 29,000 GPUs and consumed 1,287 MWh, compared to an average household which consumes 10-12MWh annually (De Vries 2023). Embodied AI agents dealing with multimodal data, especially visual data, will require more storage and computation, and consume more energy. Further, each inference request was calculated to consume as much as 3-4Wh, as much as an LED bulb running for one hour (De Vries 2023). Each planning sequence would require multiple requests and physical bodies need to carry enough computational power so that the model can be deployed in real-time.

Let us reexamine the objections of Brooks (1991) against GOFAI: "when we examine very simple level intelligence we find that explicit representations and models of the world simply get in the way. It turns out to be better to use the world as its own model." Does it apply to WEAI as well? It largely does. The WEAI representations are not explicit. However, the power of WEAI is exactly in leveraging the world models. Exploiting direct interaction with the world is in theory possible if learning in a particular robot embodiment is used, but it would essentially have to bypass the LLM.

**Outlook – the future of WEAI in Robotics**
Let us sketch possible directions of future developments around WEAI.

*WEAI for flexible visual perception and planning*

The conservative scenario is that the power and LLMs and VLMs, leveraging what can be learned from data already available on the Internet, will be exploited for robotics for flexible planning and for (passive) visual perception, along the lines of the "Natural WEAI extensions" described above. The interface to the robot will still rely on a "Robot control API" taking care of real-time interaction, like in the current models that only "reason" about the end effector position or robot center of mass.

*Making WEAI more embodied*
Some developments within WEAI naturally extend it, but at the same time have deeper implications. One such development is the use of ego-centric views in the visual inputs as in (J. Yang et al. 2024). Unlike the models that use third person views and "over-the-shoulder" cameras that see the world but also the robot in it, ego-centric views may open the door to direct or active perception and sensorimotor coordination. We also argue that when datasets are collected by teleoperating robots, the view of the robot operator and the one that is recorded for subsequent training should be an ego-centric one.

Another important concept is that of object affordances—what one can do with an object (e.g., a chair has a "sit on" affordance) (Gibson 1979). Affordances have also been discussed in WEAI works (Ahn et al. 2022; Singh et al. 2023) where they are treated as properties of the objects alone. However, affordances depend on the perceptual and motor capabilities of the agent (a robot arm cannot sit on a chair) and the ability to perceive them is acquired by the agent through a long sensorimotor learning process (Jamone et al. 2018). This perspective should find its way into the models, introducing robot-aware affordances (Schiavi et al. 2023).

*New foundation models for robotics?*
Perhaps, new foundation models will have to be created for robotics. However, that is associated with both practical and fundamental roadblocks.

The obvious practical complication is that such data is hard to come by and although several efforts are underway (for example (Walke et al. 2023); leaving cross-embodiment aside for now), the quantity of the data collected from real robots is nothing compared to the text and image data available on the internet. Data augmentation techniques or realistic robot simulators could constitute a partial remedy.

However, there are also fundamental problems with foundation models for robotics. In natural language processing, huge text corpora themselves provide all the training data that are needed—models can be trained to predict next words, next sentences or pick up also longer-range relationships thanks to the attention mechanisms in transformers (Vaswani 2017). Image databases like ImageNet (Deng et al. 2009) typically need labels (what is in the picture) in addition. Robot datasets, collected by driving a sensorized robot in an environment and teleoperating it to perform a task, would comprise of time series of multimodal data that can be used to train models. Driving around a real robot with cameras in environments with varying lighting conditions already introduces more variety in the collected images than in typical image datasets. With different robots, the variability of the multimodal datasets would grow dramatically, substantially reducing what can be transferred across the different platforms. There is thus a fundamental trade-off: the more diverse the robot embodiments and the more of their different sensory and action spaces is represented by the model, the less can the individual robots profit from the shared gigantic "brain". This is a principled limitation of foundation models for robotics and only partial workarounds exist: (1) abstraction layer hiding away the robot embodiment like in the low-dimensional discrete action space in (Padalkar et al. 2024; J. Yang et al. 2024); (2) standardize robots—in (Walke et al. 2023) only one robot type is used.

Another fundamental limitation is the passive mode of data collection (see Active embodied interaction versus offline learning). Imitation learning allows the model to copy how a human operating a robot solves a task but the situation resembles the one in Fig. 3. To exploit embodiment, active learning through closed-loop embodied interaction with the environment is necessary.

Finally, if foundation models for robotics continue focusing on the "robot brain" only like in Fig. 3C, they will still constitute a modern subsymbolic sense-think-act architecture with a single loop through the robot and environment which essentially also means a single time scale for the interaction (cf. (Iida and Giardina 2023)). To make proper use of all the implications of embodiment, multiple interaction loops at different time scales should be considered and more details about the mechanical and sensory system added to the model, such that the model can exploit them (see Fig. 1). Alternatively, these loops may not need to be re-represented (modeled—"sometimes the world is its own best model" (R. A. Brooks 1991)), but active learning will be necessary to take advantage of them.

*Active learning in a single embodiment (RL instead of LLM)*
Robot learning (Peters et al. 2016) has also relied on deep neural networks (like deep reinforcement learning) but not on foundation models. The learning has also typically been active and in a single robot embodiment, often with a big part of it done in simulation (e.g., FastGrasp'D (Turpin et al. 2023)). This approach is inherently much more embodied and will not hit the fundamental roadblocks described above.

*Touch is not like vision*
Tactile inputs are being added to some MLLMs recently. The approach that worked so well for vision can extend to touch provided that there is a unified representation for touch. Yang et al. (2024) developed UniTouch: a unified tactile model for vision-based touch sensors. However, there are several roadblocks: (1) there is "no touch on the Internet", (2) tactile data is strongly dependent on the robot hand or gripper used and the type of sensor (see also (Pliska et al. 2024)), (3) the way haptic (tactile and proprioceptive) data is collected is intrinsically active—the explorative movements are key for recognition.

*Cross-embodiment outlook*
Contradictory at first glance, the cross-embodiment idea is going to continue to receive attention. The bigger the embodiment gap, the less positive transfer there can be. However, at least in the motor domain, progress seems possible. For example, one controller for very diverse robot morphologies was developed in (Bohlinger et al. 2024). In biology, there also seems to be a common design of motion controllers across species (Ijspeert and Daley 2023). The idea of low-dimensional descending commands from the brain work in concert with low-level spinal control, feedforward and feedback, could be attractive for robotics foundation models—the foundation models would learn to supply these descending inputs (like locomotion speed and direction) and the rest will be up to low-level robot controllers.

*Truly Embodied AI*

Similarly to GOFAI, for WEAI the robot embodiment constitutes the uninteresting periphery that should be abstracted or standardized. This form of "Cartesian dualism" will in the end hinder progress. For optimal performance, co-design or co-evolution of brains and bodies will be necessary. This is of course a big challenge but there are works in this area approaching the problem from a modeling and control perspective (Pekarek 2010; Zardini et al. 2021) or Deep Evolutionary Learning (Gupta et al. 2021). Gupta et al. made agents undergo both lifetime learning and evolution simultaneously but asynchronously, more in line with how species of biological organisms actually evolve. They reproduced the Baldwin effect, which states that evolutionary selection prefers not only the morphologies with the highest fitness, but also the morphologies that can learn and adapt the quickest.

True embodiment will inevitably need to exploit physics. Physical interaction of robot bodies with the environment can be made part of the existing models through differentiable physics (de Avila Belbute-Peres et al. 2018). Using transformers for modeling physical systems may be possible (Geneva and Zabaras 2022). Alternatively, the controller should be given the chance to learn to exploit the interaction with the environment directly—without a model.

**Acknowledgements**
This work was co-funded by the European Union under the project ROBOPROX (reg. no. CZ.02.01.01/00/22_008/0004590).